\documentclass{article}
\usepackage{amsmath,epsfig}
\usepackage[preprint]{spconfa4}
\usepackage{caption} 
\usepackage[utf8]{inputenc}
\usepackage{amsfonts}
\usepackage{graphicx}
\usepackage{cite}

\copyrightnotice{978-1-6654-3864-3/21/\$31.00 ©2021 IEEE}

\let\OLDthebibliography\thebibliography
\renewcommand\thebibliography[1]{
  \OLDthebibliography{#1}
  \setlength{\parskip}{0pt}
  \setlength{\itemsep}{0pt plus 0.3ex}
}

\begin{document}\sloppy
\topmargin=0mm

\def\x{{\mathbf x}}
\def\L{{\cal L}}

\title{SPATIAL CONTENT ALIGNMENT FOR POSE TRANSFER}
%
\name{Wing-Yin Yu, Lai-Man Po, Yuzhi Zhao, Jingjing Xiong, Kin-Wai Lau}
\address{Department of Electrical Engineering, City University of Hong Kong \\ (wingyinyu8-c@my.,  eelmpo@, yzzhao2-c@my.,  jingxiong9-c@my., kinwailau6-c@my.)cityu.edu.hk}

\maketitle

\begin{abstract}
Due to unreliable geometric matching and content misalignment, most conventional pose transfer algorithms fail to generate fine-trained person images. In this paper, we propose a novel framework – Spatial Content Alignment GAN (SCA-GAN) which aims to enhance the content consistency of garment textures and the details of human characteristics. We first alleviate the spatial misalignment by transferring the edge content to the target pose in advance. Secondly, we introduce a new Content-Style DeBlk which can progressively synthesize photo-realistic person images based on the appearance features of the source image, the target pose heatmap and the prior transferred content in edge domain. We compare the proposed framework with several state-of-the-art methods to show its superiority in quantitative and qualitative analysis. Moreover, detailed ablation study results demonstrate the efficacy of our contributions. Codes are publicly available at github.com/rocketappslab/SCA-GAN.
\end{abstract}
\section{Introduction}
\label{sec:intro}

The goal of pose transfer is to synthesize a new person image based on the transformation of appearance details from a source person image and an arbitrary target pose heatmap. Due to the challenge of spatial misalignment, transferring facial characteristics and textures of garment becomes the bottlenecks of this task. It, however, contains tremendous potential multimedia applications such advertisement synthesis, movie making or data augmentation for person re-identification.

Current solutions of pose transfer can generally be divided into two types: (1) wrapped-based methods such as~\cite{dong2018soft,siarohin2018deformable} that utilizing the geometric correspondence to warp target regions according to the pose. Such wrapping can maintain style consistency of the original image but the boundaries of the human part are blurred due to unreliable geometric matching. (2) Pose-attention-based methods including~\cite{yang2020region,zhu2019progressive,huang2020generating,men2020controllable} that leveraging a series of pose-attentional blocks to encode and decode the original image according to the pose heatmap. Although this kind of methods can effectively fuse the style of original image and the target pose, due to insufficient content guidance, regions without pose landmarks especially the texture of garment lead to reconstruction distortion. Note that the RATE-Net\cite{yang2020region} attempts to solve this issue by introducing a texture enhancement module to sum the coarse output with a residual texture map. However, the enhancement is not obvious and it suffers from facial representation distortion leading to unsatisfactory visual quality.

In this paper, we propose a novel framework – Spatial Content Alignment GAN (SCA-GAN) to mainly enhance the content consistency for garment texture and human facial characteristics. We adopt a two-phase approach to map the misaligned content information to spatially aligned feature space. Firstly, we employ a Prior Content Transfer Network (PCT-Net) to transfer the edge content in a separation manner. Instead of using part-level parsing labels suggested in~\cite{dong2018soft}, we leverage pixel-level edge maps to explicitly highlight high-frequency signals which dominated a wide range of spectrum for content information. The prior transferred edge map can be served as an extra constraint in high-level perspective to provide spatial hints for the characteristics of both person identity and garments. Secondly, we develop a new Image Synthesis Network (IS-Net) to synthesize photo-realistic person images according to the appearance of source images, target pose heatmaps and the prior transferred content maps in edge domain. To encapsulate the unique pattern statistics from the original image, we exploit the residual-like Style Encoder Blocks (Style EnBlk) to strengthen the ability of feature extraction during the encoding stage. After that, we propose the Content-Style SPADE (CS-SPADE) to synthesize fine-grained appearance details by generalizing the input source for accepting both aligned content images (pose and edge) and misaligned style feature maps during the normalization. We conduct extensive experiments on the challenging DeepFashion~\cite{liu2016deepfashion} benchmark to verify the superiority of the proposed framework compared to some state-of-the-art approaches. In summary, we conclude the contributions as twofold:
\begin{itemize}
  \item We propose a new GAN-based framework – SCA-GAN to solve the spatial misalignment problem of pose transfer by generalizing the input source as spatially adaptive style and content.
  \item We address the problem of insufficient content information by leveraging edge maps as an extra constraint with pose heatmap. It guides the network to produce more photo-realistic person images through texture enhancement.
\end{itemize}

\section{Related works}

\subsection{Pose transfer}

Since PG\textsuperscript{2}~\cite{ma2017pose} initiated the task of pose transfer, it has drawn a lot of attention from the research field recently. Def-GAN~\cite{siarohin2018deformable} proposed a piece-wise deformable skip connection to wrap misaligned features based on the pose heatmap. Meanwhile, the Warping-GAN~\cite{dong2018soft} suggested to use a geometric matcher to allocate the style transformation according to the part-level correspondence. However, the wrapping causes blurry effect for the boundaries of person parts because of unreliable geometric matching. Recently, the PATN~\cite{zhu2019progressive} utilized the pose-attentional blocks to guide the deformable transfer process progressively. APS~\cite{huang2020generating} extended the mechanism of pose attention to encoder-decoder architecture while using adaptive normalization to embed the appearance representation with the pose heatmaps. RATE-Net~\cite{yang2020region} tried to perform texture enhancement on adaptive regions by adding the coarse output with a residual texture map. These methods disentangle style the representations with the excitation of pose landmarks. Due to lack of sufficient content guidance, regions without pose landmarks especially the texture of the garment often fail to be reconstructed. Even with the enhancement of\cite{yang2020region}, the person identities cannot be well persevered due to spatially misaligned entanglement.

\subsection{Image translation}

Generative Adversarial Networks (GANs)\cite{goodfellow2014generative} have shown great promise for image translation by leveraging a generator and discriminator network to minimize domain differences between the generated images and real samples. Pix2Pix\cite{isola2017image} demonstrated the effectiveness of conditional image translation based on specific input constraints. Recently, a spatially adaptive denormalization (SPADE)\cite{park2019semantic} method was proposed to inject content information by the variant of Adaptive Instance Normalization (AdaIN)\cite{huang2017arbitrary}. In this paper, we generalize the spatial hints from the aligned content domain to the misaligned style domain so as to optimize the fusion process with learnable parameters. It can maintain more style correspondence and content coherence such as the texture of garments and person facial characteristics.

\section{Proposed Method}

\begin{figure}[t]
\centering
{\includegraphics[width=\linewidth]{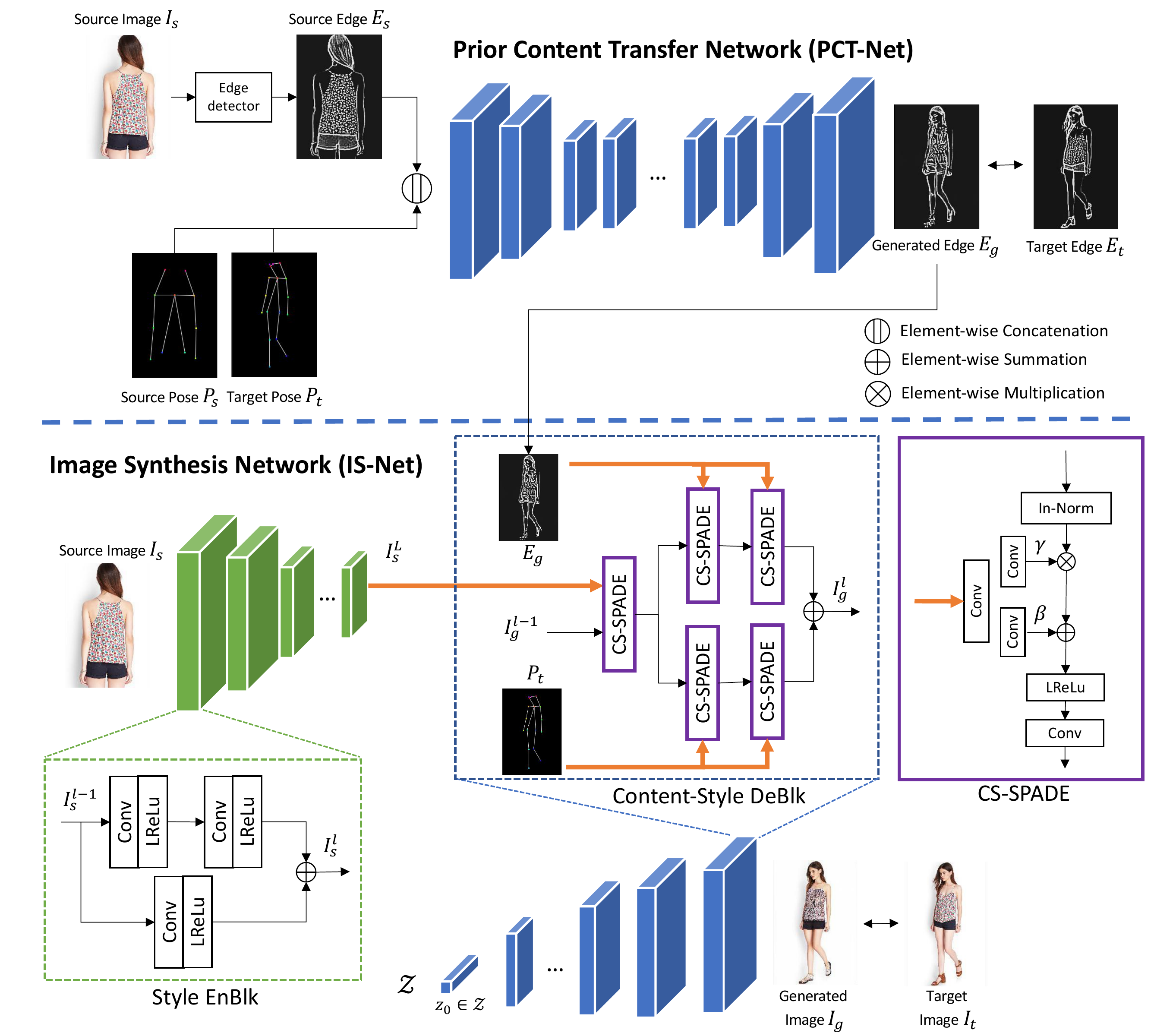}}
\caption[width=\linewidth]{The overview of our SCA-GAN. We firstly perform content transfer via the PCT-Net. We then use the transferred edge content and the target pose to provide spatial transformation guidance for style injection. The IS-Net is responsible for progressively synthesizing a realistic-looking person image using the Style EnBlk and Content-Style DeBlk.}
\label{fig:architecture}
\end{figure}

We adopt a two-phase approach to establish the whole transferring network including Phase one – Prior Content Transfer Network to deliver the content consistency and Phase two – Image Symphysis Network to combine referenced style with conditional contents. An overview of network architecture is illustrated in Figure~\ref{fig:architecture}. 

\subsection{Problem definition}

The objective of pose transfer is to generate a person image $I_g$ based on the source person image $I_s$ and the target pose heatmap $P_t$. The more appearance details $I_s$ are transferred to $P_t$, the higher similarity between $I_g$ and the target person image $I_t$ the result can get. In this paper, we define two models including Prior Content Transfer Network (PCT-Net) $\mathcal{F}(\cdot)$ and Image Symphysis Network (IS-Net) $\mathcal{G}(\cdot)$.

\subsection{Prior Content Transfer Network (PCT-Net)}

It is difficult to directly map non-aligned features from $I_s$ to $I_g$ since the feature space of $I_s$ is complex due to mixture of style and content information. In order to reduce the complexity of image synthesis process, inspired by\cite{dong2018soft}, we exploit the PCT-Net to transfer the content in the first phase. Instead of using part-level parsing labels, we suggest leveraging pixel-level edge maps to explicitly highlight the high-frequency signals which dominated a wide range of spectrum for content information. We believe that the prior transferred edge content can deliver crucial spatial hints for the characteristics of both person identity and garments. From Figure~\ref{fig:architecture}, we firstly apply an arbitrary edge detector $\mathcal{M}\left(\cdot\right)$ to perform edge detection on $I_s$. By channel-wisely concatenating the source edge $E_s=\mathcal{M}\left(I_s\right)$, $P_s$ and $P_t$ as input, we adopt Pix2Pix\cite{isola2017image} network as the baseline of the generator of PCT-Net to produce a coarse version of transferred edge content. The formulation is defined as
\begin{eqnarray}
E_g=\ \mathcal{F}\left(E_s,\ P_s,\ P_t\right).\ \ 
\end{eqnarray}

The generator of PCT-Net consists of 8 residual blocks with a maximum of 512 channels for bottlenecks. The resultant output is activated by a hyperbolic tangent function to obtain a gray-scale edge map $E_g\in(0,1)$. It will be directly fitted to Phase two – IS-Net for spatially guided symphysis purpose.

\subsection{Image Synthesis Network (IS-Net)}

In the second phase, we focus on the process of rendering a realistic-looking person image based on not only the appearance style of $I_s$ but also the spatial content of $P_t$ and $E_g$ (generated from PCT-Net). We introduce a novel GAN-based IS-Net to deal with the spatial misalignment problem by a style encoder and a content-style decoder. It is formulated as 
\begin{eqnarray}
I_g=\ \mathcal{G}\left(I_s,\ E_g,\ P_t\right).\ \ 
\end{eqnarray}

\textbf{Style encoder}. To transfer the detailed appearance from $I_s$ to $I_g$, we construct a series of encoding blocks to capture the reference style $I_s^L$. Referenced from\cite{jiang2020tsit}, the ability of feature extraction for residual blocks\cite{he2016deep} is well appreciated. As shown in Figure~\ref{fig:architecture}, we modify the standard residual blocks as our Style Encoder Block (Style EnBlk) within the style encoder. Different from previous designs, each Style EnBlk contains three convolutional layers with the same kernel size $3\times3$ following with LeakyReLU\cite{xu2015empirical} activation layers instead of normalization layers. The stride of two convolutional layers at top of the block is set to 2. The number of down-sampling layers $L$ is set to 6. The elimination of normalization layer in the style encoder is to prevent some unique texture patterns from being distorted by regularization since those specific characteristics such as person identity and garment design are hard to be generalized with few training samples.

\textbf{Spatially adaptive style-content encoder}. To synthesize fine-grained appearance details, we exploit a new spatially adaptive image translation approach based on SPADE\cite{park2019semantic} denormalization unit. We define it as Content-Style SPADE (CS-SPADE) since we generalize the input source for accepting both aligned content images and misaligned style feature maps during the normalization period. Then, we can simultaneously inject content and style information with learnable parameters to optimize the fusion process. Instead of using batch normalization\cite{ioffe2015batch}, we apply instance normalization\cite{ulyanov2016instance} to provide visual and appearance alignment. Formally, the activation features of CS-SPADE at space $(n\in N,\ c\in C^i,\ h\in H^i,\ w\in W^i)$ is:
\begin{multline}
Conv(LReLU(\\ 
\gamma_{c,h,w}^i(S)\ \cdot\ \frac{f_{n,c,h,w}^i-u_{nc}^i}{\sigma_{nc}^i}+\beta_{c,h,w}^i(S))) ,
\end{multline}
where $u_{nc}^i$ and $\sigma_{nc}^i$ are the mean and stand deviation of $f_{n,c,h,w}^i$ which is a function of either a sample of Gaussian noise $z_0$ at the first up-sampling layer or the previous synthesized feature maps $I_g^{l-1}$, i.e. $f_{n,c,h,w}(I_g^{l-1})$ in channel $c$ of mini-batch $n$: 
\begin{gather} 
u_{nc}^i=\frac{1}{H^iW^i}\sum_{h,w} f_{n,c,h,w}^i\ ,\\
\left(\sigma_{nc}^i\right)^2=\frac{1}{H^iW^i}\sum_{h,w}{f_{n,c,h,w}^i}^2-{u_{nc}^i}^2\ \ .
\end{gather}

The learnable parameters scale $\gamma_{c,h,w}^i(S)$ and bias $\beta_{c,h,w}^i(S)$  are computed by convoluting either one of the input source $S$ including the conditional style $I_s^L$ from last layer of Style Encoder, the target pose $P_t$ and the prior transferred content edge map $E_g$ from PCT-Net in Phase one. 

Finally, the spatially adaptive style-content encoder is constructed by a series of Content-Style Decoding Block (Content-Style DeBlk) from coarse-to-fine structure. The output of the Content-Style DeBlk is the fusion of pose-edge guided content representations demodulated by the conditional style representation: 
\begin{eqnarray}
I_g^l=\mathcal{U}\left(E_g,\ \mathcal{W}\left(I_g^{l-1},I_s^L\right)\right)+\mathcal{V}\left(P_t,\ \mathcal{W}\left(I_g^{l-1},I_s^L\right)\right),
\end{eqnarray}
where $\mathcal{W}(\cdot)$ is responsible for demodulating the previous generation result $I_g^{l-1}$ with the conditionally misaligned style representation $I_s^L$. $\mathcal{U}(\cdot)$ and $\mathcal{V}\left(\cdot\right)$ are two spatially adaptive stylizers to demodulate the spatially aligned content representations for producing a fine-grained and transferred visual representation.

\section{Training and Optimization}

\subsection{Objective functions}

Following similar training strategy with the existing pose transfer works\cite{zhu2019progressive,huang2020generating,men2020controllable}, the training loss is coupled from four losses including adversarial loss $\mathcal{L}_{adv}$, appearance loss $\mathcal{L}_1$, perceptual loss $\mathcal{L}_{per}$ and contextual loss $\mathcal{L}_{cx}$ as follow:
\begin{equation} \label{eqn:loss}
\mathcal{L}_{full}=\lambda_{adv}\mathcal{L}_{adv}+{\lambda_1\mathcal{L}}_1+{\lambda_{per}\mathcal{L}}_{per}+{\lambda_{cx}\mathcal{L}}_{cx}\ ,
\end{equation}
where $\lambda_{adv}$, $\lambda_1$, $\lambda_{per}$ and $\lambda_{cx}$ are corresponding to different weighting parameters to optimize the result respectively. 

\textbf{Adversarial loss.} We utilize the adversarial loss $\mathcal{L}_{adv}$ to maintain both visual style and pose content consistency by leveraging two independent discriminators $D_s$ and $D_c$. The $D_\ast$ consists of two down-sampling convolutional layers following with three residual blocks for enhancement of discriminative capability. The joint loss terms are formulated as:
\begin{align} 
&\mathcal{L}_{adv}=\nonumber \\
&\mathbb{E}_{I_s,I_t,I_g}\left[\log{\left(D_s\left(I_s,I_t\right)\right)}+\log{\left(1-D_s\left(I_s,I_g\right)\right)}\right]+\nonumber \\
&\mathbb{E}_{I_t,P_t,I_g}\left[\log{\left(D_c\left(P_t,I_t\right)\right)}+\log{\left(1-D_c\left(P_t,I_g\right)\right)}\right],
\end{align}
where $D_s$ and $D_c$ are the visual style discriminator and pose content discriminator; ${(I_s,I_t)}\in\mathbb{I}_{real}$, $P_t\in\mathbb{P}_{real}$ and $I_g\in\mathbb{I}_{fake}$ indicate samples from the distribution of real person image, real pose heatmap and generated person image. 

\textbf{Appearance loss.} To enforce discriminative loss, we employ a pixel-wise L1 loss to synthesize plausible appearance compared to the ground-truth image. 
\begin{eqnarray}
\mathcal{L}_1=\frac{1}{CHW}\sum_{c,h,w}\|{I_g|_{c,h,w}-I_t|_{c,h,w}}\|_1.
\end{eqnarray}

\textbf{Perceptual loss.} To minimize the distance in feature space, we adopt the standard perceptual loss\cite{johnson2016perceptual} in our network. It aims to enhance the visual quality by increasing the similarity of feature matching. Precisely, it computes the pixel-wise L1 difference of a selected layer $\ell=Conv1\_2$ from a pre-trained VGG-19\cite{simonyan2014very} model $\theta_\ell\left(\cdot\right)$. It is defined as:
\begin{eqnarray}
\mathcal{L}_{per}=\frac{1}{C_\ell H_\ell W_\ell}\sum_{c,h,w}\|{\theta_\ell(I_g)|_{c,h,w}-\theta_\ell(I_t)|_{c,h,w}}\|_1.
\end{eqnarray}

\textbf{Contextual loss.} To maximize the similarity between two spatially misaligned images in a similar context space, we exploit the contextual loss\cite{mechrez2018contextual} to allow spatial alignment according to contextual correspondence during the deformation process. It makes use of the normalized Cosine distance between two feature maps to measure the similarity of two misaligned features. It is formulated as:\\
\resizebox{.9\linewidth}{!}{
\begin{minipage}{\linewidth}
\begin{eqnarray}
\mathcal{L}_{cx}=-\frac{1}{C_\ell H_\ell W_\ell}\sum_{c,h,w}log\left[CX\left(\delta_\ell\left(I_g^L\right)|_{c,h,w},\delta_\ell\left(I_t\right)|_{c,h,w}\right)\right],
\end{eqnarray}
\end{minipage}
}
\\[8pt] where $l=relu\{3\_2,4\_2\}$ layers from a pre-trained VGG-19\cite{simonyan2014very} model $\delta(\cdot)$, further details of the similarity function $CX(\cdot)$ may be found in\cite{mechrez2018contextual}. 

\begin{table}[t]
\begin{center}
\caption{Quantitative analysis of comparison against the current state-of-art methods. The best scores are highlighted with bold format.}
\label{tab:sota_quan}
\begin{tabular}{l|cccc}
\hline
\multicolumn{1}{c|}{Methods} & IS$\uparrow$             & SSIM$\uparrow$           & FID$\downarrow$             & LPIPS$\downarrow$          \\ \hline \hline
PG\textsuperscript{2}\cite{ma2017pose}                          & 3.202          & 0.773          & -               & -              \\
Def-GAN\cite{siarohin2018deformable}                      & 3.362          & 0.760          & 18.457          & 0.233          \\
RATE-Net\cite{yang2020region}                     & 3.125          & 0.774          & 14.611          & 0.218          \\
PATN\cite{zhu2019progressive}                         & 3.209          & 0.773          & 19.816          & 0.253          \\
APS\cite{huang2020generating}                          & 3.295          & \textbf{0.775} & 15.017          & 0.178          \\
ADGAN\cite{men2020controllable}                        & 3.364          & 0.772          & 13.224          & 0.176          \\
SCA-GAN(Ours)                & \textbf{3.497} & \textbf{0.775} & \textbf{11.676} & \textbf{0.167} \\ \hline
\end{tabular}
\end{center}
\end{table}

\begin{figure}[t]
\centering
{\includegraphics[width=\linewidth]{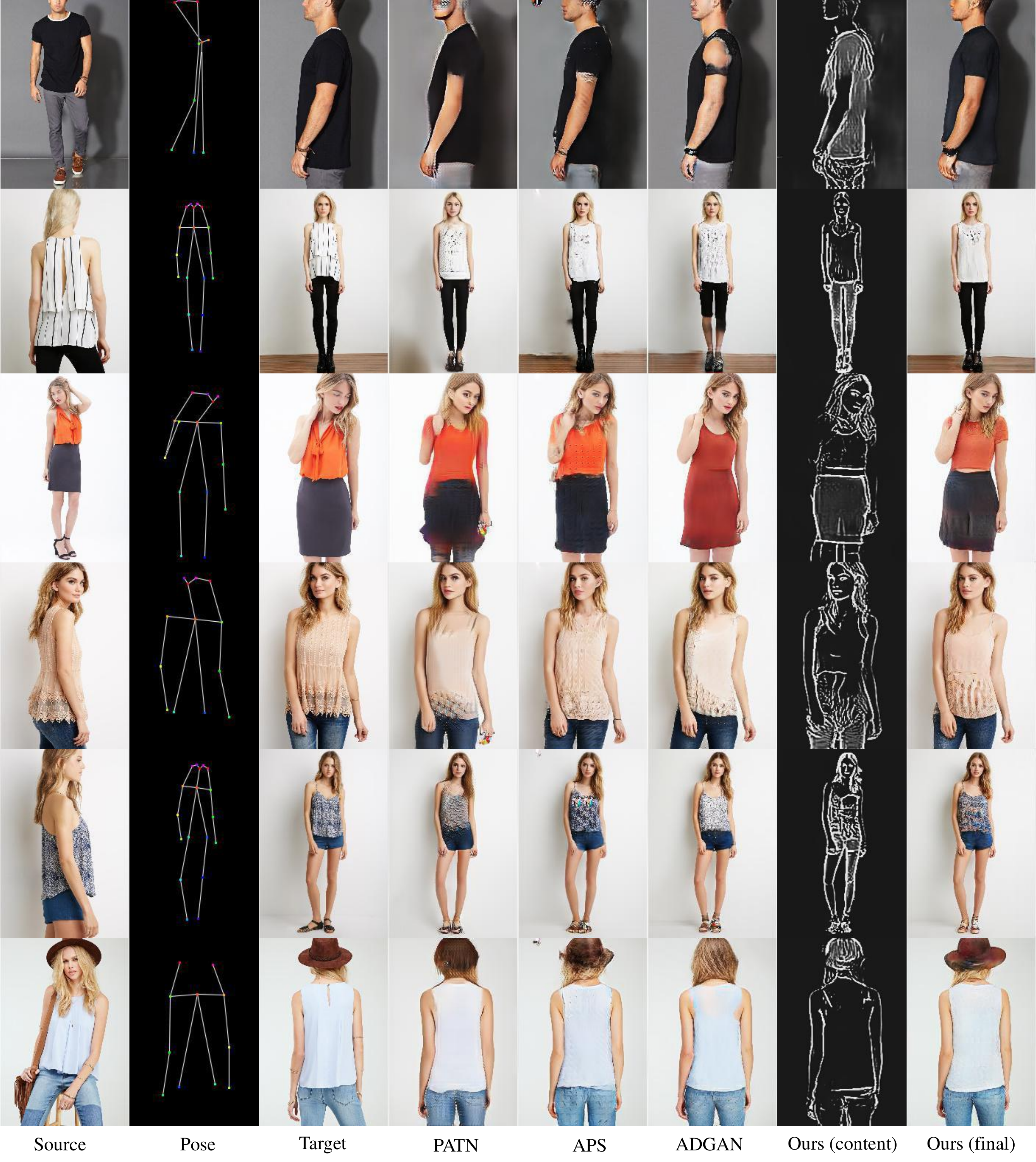}}
\caption[width=\linewidth]{Qualitative results of comparison against current state-of-art methods. Please zoom in for details.}
\label{fig:comparison}
\end{figure}

\subsection{Implementation settings}

We use the same training loss in Equation~\ref{eqn:loss} for both PCT-Net and IS-Net except switching the main input source from person images to edge maps. We use Adam optimizer\cite{kingma2014adam} with momentum 0.5 to train each model for 400 epochs. During the training of IS-Net, we use the online generated results from the pre-trained PCT-Net. The learning rate is set to 1e-4 which is linearly decayed to 0 after 200 epochs. The negative slope of LeakyReLU is set to 0.2. The weighting parameters for $\lambda_{adv}$, $\lambda_1$, $\lambda_{per}$ and $\lambda_{cx}$ are set to 5, 1, 1 and 0.1.

\section{Experimental Results}
%

\textbf{Dataset.} We conduct experiments on \textit{In-shop Clothes Retrieval Benchmark} DeepFashion\cite{liu2016deepfashion} containing high-resolution person images with different poses and appearances. There are totally 101,966 training pairs and 8,570 testing pairs. 
We found that the Extended Difference-of-Gaussians (XDoG)\cite{winnemoller2012xdog} edge detection method can preserve more content information for our prior content transfer. We use HPE\cite{cao2017realtime} as pose detector to generate the pose heatmaps. All input images are resized to $256\times176$ dimensions.

\textbf{Evaluation Metrics.} For general image generation tasks, Inception Score (IS)\cite{salimans2016improved} and Structural Similarity (SSIM)\cite{wang2004image} are two widely used evaluation metrics to quantify the perceptual performance and image quality. Following with\cite{yang2020region}, we employ two supervised perceptual metrics including Learned Perceptual Image Patch Similarity (LPIPS)\cite{zhang2018unreasonable} and Fréchet Inception Distance (FID)\cite{heusel2017gans} to consolidate the visual quality assessment in terms of perceptual distance between the generated images and real images.

\begin{table}[t]
\begin{center}
\caption{Quantitative results of ablation study. The best scores are highlighted with bold format.}
\label{tab:ablation_quan}
\resizebox{\linewidth}{!}{
\begin{tabular}{l|cccc}
\hline
\multicolumn{1}{c|}{Methods} & IS$\uparrow$	             & SSIM$\uparrow$           & FID$\downarrow$             & LPIPS$\downarrow$          \\ \hline \hline
with semantic content        & 3.388          & 0.762          & 18.634          & 0.195          \\
w/o prior transfer        & 3.365          & 0.774          & 12.402          & \textbf{0.166} \\
w/o content branch           & 3.274          & 0.756          & 19.503          & 0.194          \\
SPADE ResBlk                 & 3.343          & 0.772          & 13.888          & 0.170          \\
Batch-Norm (encoder)        & 3.340          & \textbf{0.775} & 12.197          & \textbf{0.166} \\
In-Norm (encoder)     & 3.301          & 0.773          & 12.738          & 0.167          \\
Full model                   & \textbf{3.497} & \textbf{0.775} & \textbf{11.676} & 0.167          \\ \hline
\end{tabular}}
\end{center}
\end{table}

\subsection{Comparison with state-of-the-art methods}

We evaluate the quantitative and qualitative performance with current state-of-the-art methods.

\textbf{Quantitative comparison.} As shown in Table~\ref{tab:sota_quan}, our method outperforms the current state-of-the-art methods both in unsupervised and supervised perceptual metrics. Specifically, our method achieves a significant gain, around 4\% increment, in term of IS score compared to ADGAN\cite{men2020controllable}, and even more compared to other pose-attentional approaches. We believe that it is beneficial to the edge content guidance for supporting the geometric transformation of fine-gained appearance details. The SCA-GAN also gets a comparable performance for the SSIM values which showing the robustness to generate high quality images with natural texture synthesis. Moreover, our method obtains the best FID and LPIPS scores with obvious improvement. It can certify the photo-realistic generation ability to recover misaligned style and content information with arbitrary pose transfer.     

\textbf{Qualitative comparison.} 
From Figure~\ref{fig:comparison}, we show some qualitative examples from diverse viewpoints including full body, side view, back view, upper-half view and partial view. It is clear that our generated images get the highest visual similarity compared to the target images. The prior generated content edge maps demonstrate the success of spatial content alignment for misaligned image translation. It can considerably highlight the important content information to assist the IS-Net for plausible image reconstruction. For example, the completeness of T-shirt in the first row, the natural fusion of pants with the floor in the second row, the fashion consistency of the garments in the third row, the rich crease on the dress in the fourth row are all pre-transferred from the edge content domain. The vivid appearance of camisole presented in the fifth row showcases the strong ability of style encoding and decoding. Lastly, although there is a chance that the prior content transfer fails to highlight all content information due to class-imbalance problem in DeepFashion dataset, the SCA-GAN can still distinctly synthesize important objects presented in the source image, such as the hat in the last row. It demonstrates that there is a complementary stimulation between the Style EnBlk and Content-Style DeBlk to cooperate the synthesis process. 

\subsection{Ablation study}

\begin{figure}[t]
\centering
{\includegraphics[width=\linewidth]{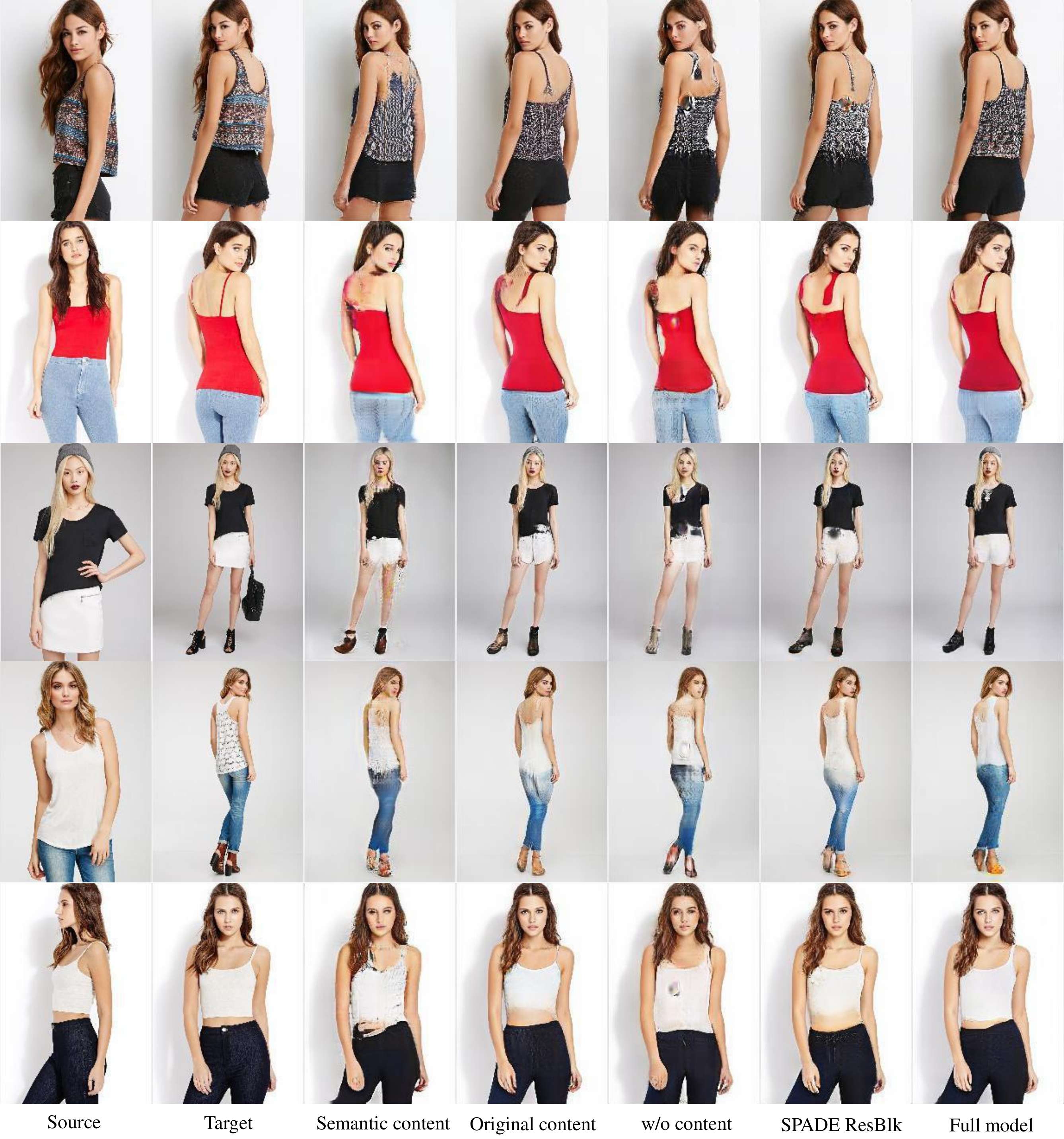}}
\caption[width=\linewidth]{Qualitative results of ablation study. Please zoom in for details.}
\label{fig:ablation}
\end{figure}

To verify the efficacy of our SCA-GAN, we show the result of ablation study as follow.

\textbf{Quantitative comparison.} As shown in Table~\ref{tab:ablation_quan}, we conduct several experiments focusing on the prior content transfer branch. Compared to semantic content (i.e. part-level segmentation map), our edge content can boost the performance for all perceptual metrics with significant increments. Without prior transfer (i.e. use edge map from source image as input) or without the whole branch, the robustness is not guaranteed since there are no sufficient conditional transformation hints to perform detailed spatial alignment. Our Content-Style DeBlk outperforms the original SPADE ResBlk with obvious improvement since it can generalize the input source for accepting both aligned content images and misaligned style features. Finally, batch normalization and instance normalization are also evaluated in our method but the improvements are trivial. It shows that our Style EnBlk can preserve a better appearance encapsulation even though there are no regularization techniques used in our style encoder. 

\textbf{Qualitative comparison.} Perceptually, the full framework can preserve more characteristics of the garments as illustrated in Figure~\ref{fig:ablation}. With the assistance of spatial content alignment, it can ensure the completeness of fashion garments while maintaining high fidelity of human facial characteristics. Moreover, it can preserve better boundary reconstruction between some ambiguous objects since the content consistency is well pre-activated. Lastly, the over-smoothing problem shown in the short pants and jeans is also addressed by the complementary fusion of style and content injection.

\section{Conclusion}

In this paper, we propose a novel SCA-GAN to generate plausible person images with pose conditions. We attempt to solve the spatial misalignment problem for this task by using a prior edge content transfer method with complementary style and content injection. Our generalization of style and content information can produce photo-realistic person images through spatial content alignment. The proposed framework can surpass the current state-of-the-art methods for both supervised and unsupervised perceptual metrics with significant improvement while it can generate much fine-grained texture of garments and characteristics of the person. 

\bibliographystyle{IEEEbib}
{\footnotesize
\bibliography{icme2021template}}

\begin{thebibliography}{10}

\bibitem{dong2018soft}
Haoye Dong, Xiaodan Liang, Ke~Gong, Hanjiang Lai, Jia Zhu, and Jian Yin,
\newblock ``Soft-gated warping-gan for pose-guided person image synthesis,''
\newblock {\em Advances in neural information processing systems}, vol. 31, pp.
  474--484, 2018.

\bibitem{siarohin2018deformable}
Aliaksandr Siarohin, Enver Sangineto, St{\'e}phane Lathuiliere, and Nicu Sebe,
\newblock ``Deformable gans for pose-based human image generation,''
\newblock in {\em Proceedings of the IEEE Conference on Computer Vision and
  Pattern Recognition}, 2018, pp. 3408--3416.

\bibitem{yang2020region}
Lingbo Yang, Pan Wang, Xinfeng Zhang, Shanshe Wang, Zhanning Gao, Peiran Ren,
  Xuansong Xie, Siwei Ma, and Wen Gao,
\newblock ``Region-adaptive texture enhancement for detailed person image
  synthesis,''
\newblock in {\em 2020 IEEE International Conference on Multimedia and Expo
  (ICME)}. IEEE, 2020, pp. 1--6.

\bibitem{zhu2019progressive}
Zhen Zhu, Tengteng Huang, Baoguang Shi, Miao Yu, Bofei Wang, and Xiang Bai,
\newblock ``Progressive pose attention transfer for person image generation,''
\newblock in {\em Proceedings of the IEEE Conference on Computer Vision and
  Pattern Recognition}, 2019, pp. 2347--2356.

\bibitem{huang2020generating}
Siyu Huang, Haoyi Xiong, Zhi-Qi Cheng, Qingzhong Wang, Xingran Zhou, Bihan Wen,
  Jun Huan, and Dejing Dou,
\newblock ``Generating person images with appearance-aware pose stylizer,''
\newblock {\em arXiv preprint arXiv:2007.09077}, 2020.

\bibitem{men2020controllable}
Yifang Men, Yiming Mao, Yuning Jiang, Wei-Ying Ma, and Zhouhui Lian,
\newblock ``Controllable person image synthesis with attribute-decomposed
  gan,''
\newblock in {\em Proceedings of the IEEE/CVF Conference on Computer Vision and
  Pattern Recognition}, 2020, pp. 5084--5093.

\bibitem{liu2016deepfashion}
Ziwei Liu, Ping Luo, Shi Qiu, Xiaogang Wang, and Xiaoou Tang,
\newblock ``Deepfashion: Powering robust clothes recognition and retrieval with
  rich annotations,''
\newblock in {\em Proceedings of the IEEE conference on computer vision and
  pattern recognition}, 2016, pp. 1096--1104.

\bibitem{ma2017pose}
Liqian Ma, Xu~Jia, Qianru Sun, Bernt Schiele, Tinne Tuytelaars, and Luc
  Van~Gool,
\newblock ``Pose guided person image generation,''
\newblock in {\em Advances in neural information processing systems}, 2017, pp.
  406--416.

\bibitem{goodfellow2014generative}
Ian Goodfellow, Jean Pouget-Abadie, Mehdi Mirza, Bing Xu, David Warde-Farley,
  Sherjil Ozair, Aaron Courville, and Yoshua Bengio,
\newblock ``Generative adversarial nets,''
\newblock in {\em Advances in neural information processing systems}, 2014, pp.
  2672--2680.

\bibitem{isola2017image}
Phillip Isola, Jun-Yan Zhu, Tinghui Zhou, and Alexei~A Efros,
\newblock ``Image-to-image translation with conditional adversarial networks,''
\newblock in {\em Proceedings of the IEEE conference on computer vision and
  pattern recognition}, 2017, pp. 1125--1134.

\bibitem{park2019semantic}
Taesung Park, Ming-Yu Liu, Ting-Chun Wang, and Jun-Yan Zhu,
\newblock ``Semantic image synthesis with spatially-adaptive normalization,''
\newblock in {\em Proceedings of the IEEE Conference on Computer Vision and
  Pattern Recognition}, 2019, pp. 2337--2346.

\bibitem{huang2017arbitrary}
Xun Huang and Serge Belongie,
\newblock ``Arbitrary style transfer in real-time with adaptive instance
  normalization,''
\newblock in {\em Proceedings of the IEEE International Conference on Computer
  Vision}, 2017, pp. 1501--1510.

\bibitem{jiang2020tsit}
Liming Jiang, Changxu Zhang, Mingyang Huang, Chunxiao Liu, Jianping Shi, and
  Chen~Change Loy,
\newblock ``Tsit: A simple and versatile framework for image-to-image
  translation,''
\newblock {\em arXiv preprint arXiv:2007.12072}, 2020.

\bibitem{he2016deep}
Kaiming He, Xiangyu Zhang, Shaoqing Ren, and Jian Sun,
\newblock ``Deep residual learning for image recognition,''
\newblock in {\em Proceedings of the IEEE conference on computer vision and
  pattern recognition}, 2016, pp. 770--778.

\bibitem{xu2015empirical}
Bing Xu, Naiyan Wang, Tianqi Chen, and Mu~Li,
\newblock ``Empirical evaluation of rectified activations in convolutional
  network,''
\newblock {\em arXiv preprint arXiv:1505.00853}, 2015.

\bibitem{ioffe2015batch}
Sergey Ioffe and Christian Szegedy,
\newblock ``Batch normalization: Accelerating deep network training by reducing
  internal covariate shift,''
\newblock {\em arXiv preprint arXiv:1502.03167}, 2015.

\bibitem{ulyanov2016instance}
Dmitry Ulyanov, Andrea Vedaldi, and Victor Lempitsky,
\newblock ``Instance normalization: The missing ingredient for fast
  stylization,''
\newblock {\em arXiv preprint arXiv:1607.08022}, 2016.

\bibitem{johnson2016perceptual}
Justin Johnson, Alexandre Alahi, and Li~Fei-Fei,
\newblock ``Perceptual losses for real-time style transfer and
  super-resolution,''
\newblock in {\em European conference on computer vision}. Springer, 2016, pp.
  694--711.

\bibitem{simonyan2014very}
Karen Simonyan and Andrew Zisserman,
\newblock ``Very deep convolutional networks for large-scale image
  recognition,''
\newblock {\em arXiv preprint arXiv:1409.1556}, 2014.

\bibitem{mechrez2018contextual}
Roey Mechrez, Itamar Talmi, and Lihi Zelnik-Manor,
\newblock ``The contextual loss for image transformation with non-aligned
  data,''
\newblock in {\em Proceedings of the European Conference on Computer Vision
  (ECCV)}, 2018, pp. 768--783.

\bibitem{kingma2014adam}
Diederik~P Kingma and Jimmy Ba,
\newblock ``Adam: A method for stochastic optimization,''
\newblock {\em arXiv preprint arXiv:1412.6980}, 2014.

\bibitem{winnemoller2012xdog}
Holger Winnem{\"o}ller, Jan~Eric Kyprianidis, and Sven~C Olsen,
\newblock ``Xdog: an extended difference-of-gaussians compendium including
  advanced image stylization,''
\newblock {\em Computers \& Graphics}, vol. 36, no. 6, pp. 740--753, 2012.

\bibitem{cao2017realtime}
Zhe Cao, Tomas Simon, Shih-En Wei, and Yaser Sheikh,
\newblock ``Realtime multi-person 2d pose estimation using part affinity
  fields,''
\newblock in {\em Proceedings of the IEEE conference on computer vision and
  pattern recognition}, 2017, pp. 7291--7299.

\bibitem{salimans2016improved}
Tim Salimans, Ian Goodfellow, Wojciech Zaremba, Vicki Cheung, Alec Radford, and
  Xi~Chen,
\newblock ``Improved techniques for training gans,''
\newblock {\em Advances in neural information processing systems}, vol. 29, pp.
  2234--2242, 2016.

\bibitem{wang2004image}
Zhou Wang, Alan~C Bovik, Hamid~R Sheikh, and Eero~P Simoncelli,
\newblock ``Image quality assessment: from error visibility to structural
  similarity,''
\newblock {\em IEEE transactions on image processing}, vol. 13, no. 4, pp.
  600--612, 2004.

\bibitem{zhang2018unreasonable}
Richard Zhang, Phillip Isola, Alexei~A Efros, Eli Shechtman, and Oliver Wang,
\newblock ``The unreasonable effectiveness of deep features as a perceptual
  metric,''
\newblock in {\em Proceedings of the IEEE conference on computer vision and
  pattern recognition}, 2018, pp. 586--595.

\bibitem{heusel2017gans}
Martin Heusel, Hubert Ramsauer, Thomas Unterthiner, Bernhard Nessler, and Sepp
  Hochreiter,
\newblock ``Gans trained by a two time-scale update rule converge to a local
  nash equilibrium,''
\newblock {\em Advances in neural information processing systems}, vol. 30, pp.
  6626--6637, 2017.

\end{thebibliography}

\end{document}